\documentclass[letterpaper]{article} 
\usepackage{aaai2026}  
\usepackage{times}  
\usepackage{helvet}  
\usepackage{courier}  
\usepackage[hyphens]{url}  
\usepackage{graphicx} 
\usepackage{natbib}  
\usepackage{caption} 
\usepackage{algorithm}
\usepackage{algorithmic}

\usepackage{newfloat}
\usepackage{listings}
\DeclareCaptionStyle{ruled}{labelfont=normalfont,labelsep=colon,strut=off} 
\floatstyle{ruled}
\newfloat{listing}{tb}{lst}{}
\floatname{listing}{Listing}
\newcommand{\cmark}{\ding{51}}%
\newcommand{\xmark}{\ding{55}}%
\usepackage{nicematrix}
\usepackage{pifont}%
\usepackage{multirow}
\usepackage{cuted}
\usepackage{capt-of}
\usepackage{cleveref}
\usepackage{comment}

\title{Open-World Object Counting in Videos}
\author{
    Niki Amini-Naieni, 
    Andrew Zisserman
}
\affiliations{
    Visual Geometry Group, Dept.\ of Engineering Science, University of Oxford, UK\\
    nikian@robots.ox.ac.uk, az@robots.ox.ac.uk
}

\newcommand{\methodName}{\textsc{CountVid}}
\newcommand{\datasetName}{\textsc{VideoCount}}
\newcommand{\countgdbox}{\textsc{CountGD-Box}}
\newcommand{\az}[1]{{\textcolor{blue}{[az: #1]}}}
\begin{document}

\maketitle

\begin{strip}\centering
\includegraphics[width=0.88 \textwidth]{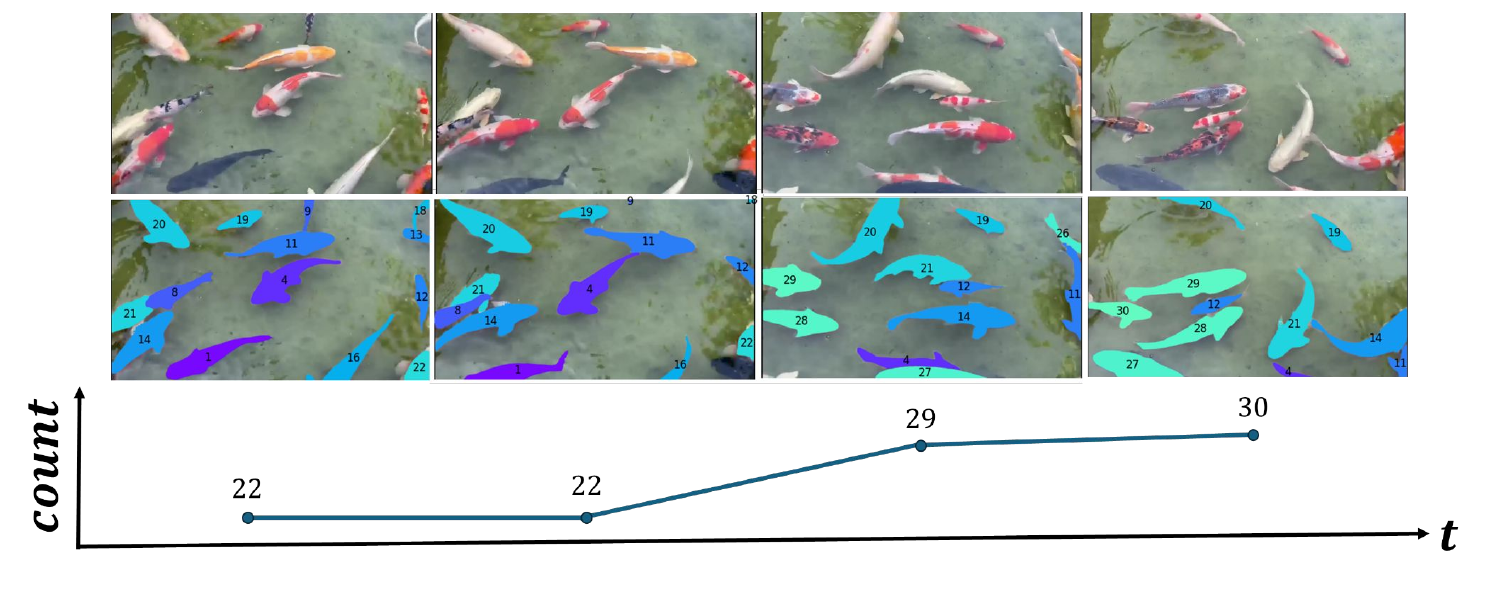}
 
\captionof{figure}{Object counting in a video. Given the video in the top row (shown by a few sample frames) and the text ``fish", \methodName\ is able to accurately match instances across the video (indicated by the color and number assigned to each fish), identify new objects, and estimate the count. The video is available on the project website.
\label{fig:teaser}}
\end{strip}

\begin{abstract}
We introduce a new task of open-world object counting in videos: given a text description, or an image example, that specifies the target object, the objective is to enumerate all the unique instances of the target objects in the video. This task is especially challenging in crowded scenes with occlusions and objects of similar appearance, where avoiding double counting and identifying reappearances is crucial. To this end, we make the following contributions: we introduce a model, \methodName, for this task. It leverages an image-based counting model, and a promptable video segmentation and tracking model, to enable automated open-world object counting across video frames. To evaluate its performance, we introduce \datasetName, a new dataset for this novel task built from the TAO and MOT20 tracking datasets, as well as from videos of penguins and metal alloy crystallization captured by x-rays. Using this dataset, we demonstrate that \methodName\ provides accurate object counts, and significantly outperforms strong baselines\footnote{\methodName\ code and the \datasetName\ dataset are available at https://www.robots.ox.ac.uk/vgg/research/countvid/.}. 
\end{abstract}

\section{Introduction}
\label{sec:intro}
Our objective in this paper is {\em open-world} object counting in videos – determining how many instances of an object class are present in a video, where the object class of interest is specified by a text description or an image exemplar and may not have been included in the training data. This is a time-dependent task, as both the current (visible count) can be reported at the frame level, or other temporal intervals, as well as the cumulative count over the entire video.

As \cref{fig:teaser} illustrates, keeping count in videos is naturally a {\em correspondence or tracking task} -- since we do not want to count the same instance multiple times, we must establish that instances in consecutive frames are the same. However, as the figure also illustrates, one of the fundamental challenges of counting in videos is {\em instance identification} -- is an object appearing in a frame a new instance? Or, is it one that drifted out of frame earlier in the sequence or was temporarily occluded? This challenge is exacerbated as the objects become indistinguishable: discriminating individual fish is possible if they have different colors and markings, but discriminating insects or crows may be impossible. 

Somewhat surprisingly, automated counting in {\em videos} is a relatively unexplored area. 
While there are a few automated methods for closed-vocabulary counting~\cite{mall_dataset, fang2019locality, count_animals, video_object_counting_dataset, han2022drvic, dronecrowd_cvpr2021, weak_vic}, there are
no open-vocabulary methods that we are aware of. This in contrast to the extensive exploration of counting in {\em images}, where open-vocabulary methods are able to use both text and exemplars to specify the target object, and can count up to thousands of instances~\cite{countgd, dave, m_Ranjan-etal-CVPR21}. Even some large-scale Vision-Language Models, such as Molmo~\cite{molmo}, are now able to accurately count beyond ten instances in an image.

This lack of research in video counting is especially surprising given the wide variety of science applications and need. Conservationists need to count animals in video sequences captured by drones for population monitoring~\cite{manual_anno_drone, Mustafa2019DETECTINGAS}. This can take up to 30 hours for a trained human analyst to annotate manually for a single one-hour flight~\cite{drones_for_conservation}. 
Material scientists count crystals forming from liquid metal alloys to determine how cooling affects the speed of the formation process~\cite{Liotti18}. Epidemiologists use human and vehicle counts from videos captured on city streets to study causes of pedestrian exposure to air pollution and mitigate them~\cite{air_pollution1, air_pollution2}. An `open-world' method that can be quickly applied to all these problems out-of-the-box, with no manual counting or additional training needed, has the potential to catalyze these applications, eliminating the annotation time, and significantly benefiting their research.

In this paper, we introduce a model, \methodName, for open-world object counting in videos that accepts a video and a prompt that specifies the target object to count as inputs, and outputs how many unique instances of the object appear in the video. The prompt can consist of a free-form text description and/or any number of `visual exemplars', where the visual exemplars indicate the object of interest by bounding boxes and can come from a video frame or an external image.

The \methodName\ model builds on advances for two distinct tasks: (i) powerful open-vocabulary image counting and detection models~\cite{countgd, Pelhan_2024_NeurIPS, zhizhong2024point}; and (ii) powerful class-agnostic video segmentation and tracking models~\cite{ravi2024sam2, yang2024samurai}. However, naively combining an image detector to provide instances for a tracker is not sufficient since state-of-the-art object detectors struggle to count high numbers of objects in densely packed scenes with many occlusions and overlapping instances~\cite{countgd, groundingrec}. 
To overcome this detection problem, for our model we leverage accurate image-based \emph{counters} that also output the bounding boxes required by the tracker~\cite{Pelhan_2024_NeurIPS, zhizhong2024point}.

One of the innovations of this paper is to 
extend the most flexible image-based counter, CountGD~\cite{countgd}, to also produce bounding boxes, offering versatility in accepting text, visual exemplar, or combined prompts and detection capabilities within the same model. We show that its performance surpasses image detectors when there are many instances of similar objects in the image, and
also prior models~\cite{countgd, dave, 10.1007/978-3-031-20044-1_20}. 

A second innovation is to propose a temporal filter to remove false positive tracks resulting from erroneous detections. The extended CountGD model, dubbed \countgdbox, and other detection-based counters are used to provide box prompts over multiple frames in the video, and the tracker is used to {\em associate} the resultant segmentations and propagate them to other frames.

To assess the performance of video counting, we introduce \datasetName, a new dataset with ground truth for this task. \datasetName\ has two types of benchmarks: first, we re-purpose standard tracking datasets, TAO~\cite{tao} and MOT20~\cite{mot20},
by adding additional annotations to ensure all objects are counted (since tracking benchmarks typically only evaluate on a subset of the objects, e.g.\ not accounting for static objects); and second, we introduce two science applications of counting with new videos containing real footage from monitoring penguins in their natural habitats and x-ray images of the crystallization process for metal alloys. The number of objects in videos from our dataset ranges from one to over a thousand.

In summary, we make the following four contributions: \emph{first} we present the novel task of open-world object counting in videos; \emph{second} we propose a model for this task, \methodName, by re-purposing and combining open-vocabulary image counting and class-agnostic segmentation and tracking models; \emph{third} we extend CountGD to produce bounding boxes as outputs, and introduce an automated method to remove false tracks; and \emph{fourth} we release \datasetName, a new video dataset for evaluating algorithms for this open-world object counting task.

\section{Related Work}
\label{sec:related_work}
A  note about terminology: in the object counting literature~\cite{AminiNaieni23, countgd, Liu22}, {\em open-world counting} refers to counting instances of an object class specified at test time via text or visual prompts. We adopt this definition, but also discuss how `open-world' has been interpreted differently in the past in the appendix. 

\paragraph{Open-World Object Counting in Images.}
Prior work on open-world object counting only focuses on images. The first image-based methods required the user to manually annotate a few example objects with `visual exemplars' to count at inference time~\cite{Liu22, Lu18, 10.1007/978-3-031-20044-1_20, m_Ranjan-etal-CVPR21, Shi2022RepresentCA, yangClassagnosticFewshotObject2021,  You_2023_WACV, low_shot, Lin_2022_BMVC}. More recent works~\cite{groundingrec, AminiNaieni23, countgd, Jiang2023CLIPCountTT, kang2024vlcounter} have leveraged pre-trained vision-language foundation models to enable the category to be specified by text. CountGD is a state-of-the-art open-world counting model that uses the joint vision-language embedding space of the Grounding DINO~\cite{liu2023grounding} foundation model to allow the user to specify the object to count with text. In addition to this, unlike most prior approaches that either only accept text or only accept visual exemplars, CountGD allows for \emph{both} inputs. By accepting only text, CountGD can adapt to novel classes without human intervention, and by accepting visual exemplars, it provides greater accuracy. We build on CountGD in this work.


\paragraph{Open-World Object Counting in Videos.}
While there is no prior work that explicitly focuses on open-world object counting in videos, there are open-world trackers~\cite{ovtrack} that can be repurposed as counters. State-of-the-art open-world trackers rely on object detectors~\cite{slack, masa, vovtrack, video_owl}. For example, the open-world tracker MASA~\cite{masa} leverages detectors such as Grounding DINO and Detic~\cite{zhou2022detecting} to first detect any object using text and then associate it across video frames. The unique objects identified and tracked throughout the video can be enumerated to estimate the count. VOVTrack~\cite{vovtrack} uses region proposals from Faster R-CNN~\cite{faster-rcnn} for object localization. Because these approaches extend detectors, they inherit their limitations.

Trackers that do not rely on pre-trained image-based object detectors also have limitations. Trackformer~\cite{trackformer} cannot adapt to novel categories at inference, only tracking objects that it has been trained to track. On the other hand, OVTR~\cite{ovtr} is an end-to-end transformer-based tracking model that can track novel objects given text input. However, it does not accept visual prompts limiting its accuracy. Furthermore, it has not been trained or tested on scenes with hundreds to over a thousand objects. SAM~2, SAM~2.1~\cite{ravi2024sam2} and SAMURAI~\cite{yang2024samurai} are recent state-of-the-art tracking and segmentation models that also adapt to novel objects without retraining. SAM~2.1 and SAMURAI extend SAM~2 with motion cues, longer video training, and occlusion handling, but both require manual prompting, which \countgdbox\ and \methodName\ automate. SAMURAI also focuses mainly on single-object tracking.

\section{\methodName\ and \countgdbox\ Models}
\label{sec:method}

\begin{figure*}[h!]
  \centering
  \includegraphics[width=0.9\linewidth]{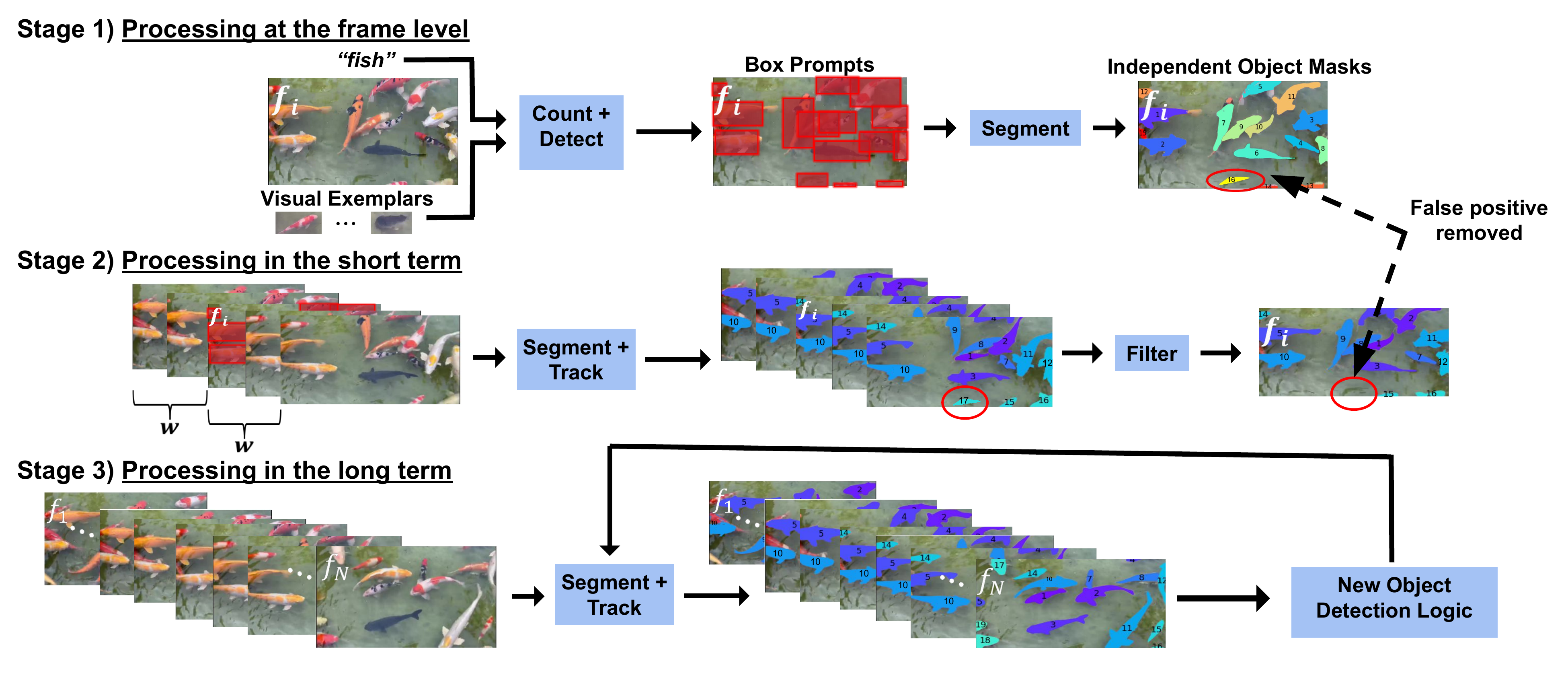}
   
  \caption{Inference with \methodName. Each stage processes the video at decreasing granularity. In Stage~1, a detection-based counter uses text and exemplars to generate box prompts for each frame, which are used by a segmentation model to produce object masks. Stage~2 applies a temporal filter over sequences of $w$ frames to remove transient false positives (e.g., the red-circled object in $f_i$). Stage~3 propagates object masks from Stage~2 across the full video while checking for new objects in each frame. All the identified objects are enumerated to produce the final global count.}
  \label{fig:method}
\end{figure*}

In this section, we first present \methodName, our method for open-world object counting in videos. We then present  \countgdbox, a multi-modal counting model that outputs bounding boxes and extends CountGD. 



\subsection{\methodName}

\methodName\ is a model that inputs a video and flexible prompts including text only, visual exemplars only, or both and outputs frame-level counts and a global count indicating the number of unique objects in the video that match the prompts. At inference, \methodName\ processes the video in three stages, at decreasing levels of granularity. The stages are illustrated in \cref{fig:method}. 


\noindent\textbf{Stage~1 -- Processing at the Frame Level.} The first stage is where the visual exemplar and text prompts are used, and the objective is to automatically obtain bounding boxes and segmentation masks for all instances of the target object in each frame. To achieve this, the visual exemplar and text prompts are fed to a counting and detection model that is applied to each video frame independently to obtain bounding boxes. The bounding boxes output by the counting model are used as box prompts for a segmentation model, which outputs masks for all the objects in the frame. We use exemplars from a single frame and apply them across the video to reduce user annotation effort. To improve efficiency, frames are subsampled before this stage begins. 

\noindent\textbf{Stage~2 -- Processing in the Short Term.} Although counting models are very accurate at counting, they can still produce false positive detections. This may occur due to motion blur. The objective of Stage~2 is to remove these false positive predictions with a temporal filter. This filter leverages the observation that false positives tend to be transient in independent  (per-frame) predictions from Stage~1, disappearing almost immediately in subsequent frames. For each detection from Stage~1, the filter checks if the object exists within a temporal window of $w$ frames. Using a segmentation and tracking model, the filter tracks $w - 1$ frames backwards and $w - 1$ frames forward from the frame the detection is in, resulting in a span of $2w-1$ frames. Objects are matched using the intersection over union (IoU) of the propagated masks from tracking and the masks from independent per-frame detections in Stage~1. An IoU greater than 0.5 is considered a match. If the object is matched in a sequence of at least $w$ consecutive frames, it is kept. This ensures the filter tolerates a degree of occlusion or intermittent visibility. Otherwise, it is removed before Stage~3 begins. Note, it is necessary to track both forwards and backwards in time, as new objects can appear (and are verified by tracking forwards in time), and also objects can disappear (e.g.\ by occlusion) and are verified by tracking backwards in time. 
 
\noindent\textbf{Stage~3 -- Processing in the Long Term.}
In the last stage, \methodName\ applies a segmentation and tracking model to the full video, keeping track of objects in the long term while checking for new objects in each frame. For each object, \methodName\ predicts a \emph{masklet}, an object mask propagated over time. New objects are detected by comparing existing masklets with per-frame masks from Stage~2. Per-frame masks that do not overlap with the existing masklets are identified as new objects. The new objects are then also tracked going forward. Once all the frames have been checked for new objects, the masklets are enumerated to estimate the final global count. The New Object Detection Logic is visually explained in detail in the appendix. 

\noindent\textbf{\methodName\ Implementation.}
We implement \methodName\ with \countgdbox\ (described below) as the counting/detection model, and SAM~2.1 as the tracker. The boxes from \countgdbox\ are used as box prompts for SAM~2.1 in Stage~1, and then SAM~2.1 tracks the masks from the prompted objects, producing masklets. Though \methodName\ is agnostic to the choice of counter/detector and tracker. We compare to other choices in the experiments.

\subsection{\textbf{\countgdbox}}\label{cgdbox}

To automatically obtain box prompts for the segmentation model, we need a detector that can handle dense scenes with many similar overlapping objects, because this will occur in the video for our challenging task. As shown by our results and prior work~\cite{countgd, AminiNaieni23}, there are open-world detectors, but they do not perform well in this setting. On the other hand, there are open-world counters that do. Out of all these counters, CountGD is the most flexible, accepting either text only, visual exemplars only, or both simultaneously to specify the object. It also provides generally strong counting performance across all prompt settings as shown by our results. However, unlike other less flexible detection-based counting models, CountGD outputs points, not boxes.

Point prompts do not specify the object to count in a well-defined way. For example, a point on the window of a car could mean every window or every car should be counted. Given there is likely more than one window on a car, this problem could result in an erroneously high or low count. This ambiguity is resolved if an image region (a box) is specified instead of a point.



To obtain well-defined object prompts for the segmentation model, we train CountGD to output \emph{boxes} in addition to \emph{points}. CountGD originally lacks this capability due to the limited bounding box data available in object counting datasets such as FSC-147~\cite{m_Ranjan-etal-CVPR21}. This scarcity exists because labeling hundreds to thousands of objects with bounding boxes is extremely tedious. Instead, these datasets only provide box annotations for a few objects per image. Drawing inspiration from DAVE~\cite{dave}, we extend CountGD to take advantage of these weak training labels. Because CountGD is built on top of the Grounding DINO architecture, it already outputs four parameters per object. The first two are used as the center of the object, while the last two are discarded by CountGD. We add two new terms, $\mathcal{L}^{e}_{h, w}$ and $\mathcal{L}^{e}_{GIoU}$, to CountGD's loss, to train the last two parameters to be the height and width of the bounding box. The new loss is defined as $\mathcal{L} = \lambda_{loc} \left(\mathcal{L}^{e}_{h, w} + \mathcal{L}_{center}\right) + \lambda_{GIoU} \mathcal{L}^{e}_{GIoU} + \lambda_{cls} \mathcal{L}_{cls}$ where $\mathcal{L}^{e}_{h, w}$ and $\mathcal{L}^{e}_{GIoU}$ are based on the bounding box regression losses from Grounding DINO. The difference here is that these losses are only calculated for the exemplars, while in Grounding DINO, they are calculated for all the objects in the image. $\mathcal{L}^{e}_{h, w}$ is the sum of the absolute errors of the height and widths and $\mathcal{L}^{e}_{GIoU}$ is the generalized intersection over union between the predicted and ground truth exemplar boxes. By training on the exemplar boxes, CountGD learns to not only predict points, but to also predict bounding boxes. We name the extended CountGD as \countgdbox\ and use it at inference to produce box prompts for the segmentation model.

\section{\datasetName: A New Counting Dataset}
\label{sec:datasets}

Current benchmarks for object counting are not sufficient for the open-world object counting in videos task. This is because existing counting datasets either only support images~\cite{m_Ranjan-etal-CVPR21, Hsieh2017DroneBasedOC} or only include a limited number of categories~\cite{mall_dataset, visdrone, mot20}. Furthermore, existing tracking datasets such as TAO~\cite{tao} only provide exhaustive annotations for a subset of objects and only label at most ten objects per video, which is far too low for practical counting use cases. Therefore, in this section, we present \datasetName, a new dataset for open-world object counting in videos that overcomes these limitations.  \datasetName\ is made up of three benchmarks: TAO-Count, MOT20-Count, and Science-Count. It contains 370 videos covering a wide range of object categories and counts as shown in \cref{our_datasets_tbl}. We include further details in the appendix. 

\begin{table}
\begin{center}
{\fontsize{9}{11}\selectfont\begin{NiceTabular}{c|c|c|c|c} 
    & \# Videos & \# Classes & \# Objects & Video Len \\
   \hline
   TAO- & \multirow{2}{*}{357} & \multirow{2}{*}{139} & 1-10 & 8s-78s \\
   Count & & & 2.7 Avg & 34s Avg\\
   \hline
   MOT20- & \multirow{2}{*}{3} & \multirow{2}{*}{1} & 80-1203 & 17s-133s \\
   Count & & & 520.3 Avg & 87s Avg\\
   \hline
   Science- & \multirow{2}{*}{10} & \multirow{2}{*}{2} & 10-154 & 4s-30s \\
   Count & & & 63 Avg & 11s Avg\\
\hline
\end{NiceTabular}}
\caption{\label{our_datasets_tbl} \datasetName\ details. Our dataset, composed of 3 benchmarks, covers a wide range of object categories (141 in total) and a wide range of object counts (1-1203 per video).}
\end{center}
\end{table}

\begin{figure*} 
  \centering
  \includegraphics[width=0.75\linewidth]{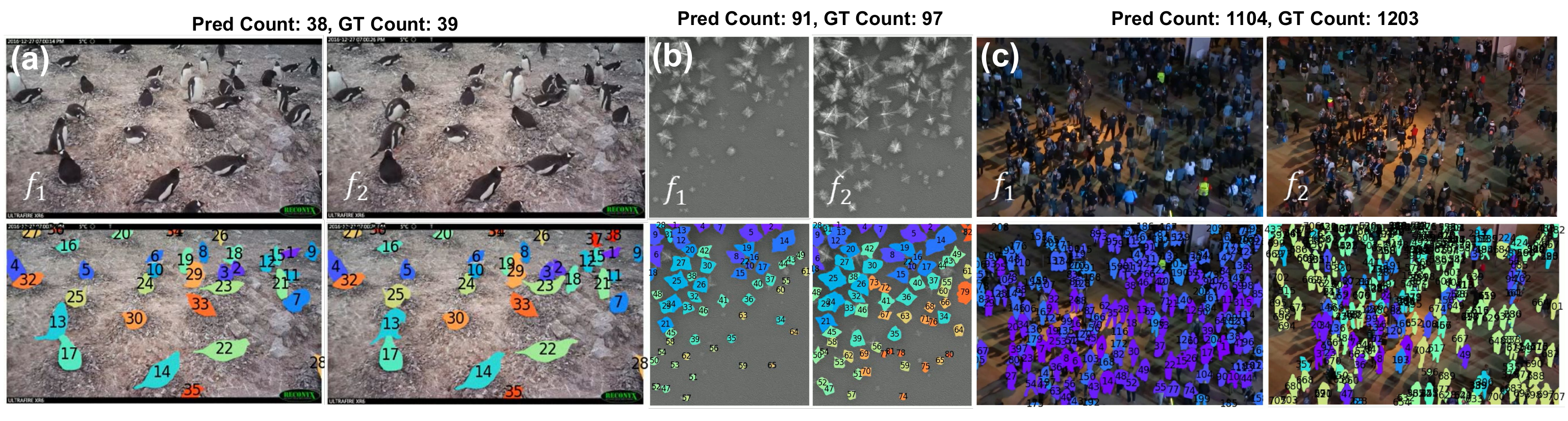}
  
  \caption{Qualitative results on \datasetName.  $f_1$ and $f_2$ are frames sampled consecutively in time. \methodName\ handles dense (b, c), deforming (b), and similar (a, b) objects.}
  \label{fig:qualitative_results}
\end{figure*}

Our dataset is built from diverse sources. For TAO-Count and MOT20-Count, we add metadata to subsets of the existing tracking datasets TAO~\cite{tao} and MOT20~\cite{mot20} specifying the counts of target objects. For Science-Count, we release new videos and count annotations from monitoring penguin populations and videoing the formation of crystals from liquid metal alloys captured with x-ray radiography~\cite{Liotti18}. Example video frames from \datasetName\ are shown in \cref{fig:qualitative_results} and the appendix. 

\datasetName\ tests \methodName's ability to adapt to a wide range of challenging scenarios. TAO-Count tests how well \methodName\ counts low numbers of objects in scenes with significant motion. MOT20-Count tests how well \methodName\ counts in heavily crowded scenes (e.g., $> 1000$ objects) with many overlapping instances. Science-Count evaluates \methodName\ on tricky real-world applications with many similar objects, some even rapidly changing structure over time in x-ray videos typically out-of-domain for foundation models. 

\section{Experiments}\label{sec:experiments}

\begin{table*}
 
\begin{center}
{\fontsize{9}{11}\selectfont\begin{NiceTabular}{l|c|c|c|c|c|c|c|c|c|c} 
 & \multicolumn{2}{c}{TAO-Count} & \multicolumn{2}{c}{TAO-Count-FSC} & \multicolumn{2}{c}{MOT20-Count} & \multicolumn{2}{c}{Penguins} & \multicolumn{2}{c}{Crystals}\\
\hline
   Method & MAE $\downarrow$ & RMSE $\downarrow$ & MAE $\downarrow$ & RMSE $\downarrow$ & MAE $\downarrow$ & RMSE $\downarrow$ & MAE $\downarrow$ & RMSE $\downarrow$ & MAE $\downarrow$ & RMSE $\downarrow$\\
   \hline
   DR. VIC (Baseline) & - & - & - & - & 100.0 & 110.8 & - & - & - & -\\
   MASA (Baseline) & 14.1 & 21.5 & 13.9 & 20.4 & 630.0 & 725.4 & 9.0 & 11.5 & 72.3 & 87.9\\
   \textbf{\methodName\ (Ours)} & \textbf{2.6} & \textbf{6.0} & \textbf{2.5} & \textbf{5.2} & \textbf{50.0} & \textbf{61.3} & \textbf{4.0} & \textbf{5.3} & \textbf{69.1} & \textbf{86.0}\\
\hline
\end{NiceTabular}}
\captionof{table}{\label{main_results_text}{Counting performance on \datasetName. MASA and \methodName\ are given \emph{\underline{only text}} prompts.} 
DR.\ VIC only counts humans and so can only be evaluated on MOT20-Count.
TAO-Count-FSC is TAO-Count with the training classes in FSC-147 removed. The class overlap between TAO-Count and FSC-147 has no significant influence on \methodName's counting accuracy.}
\end{center}
\end{table*}
 
\noindent \textbf{Implementation Details: } \countgdbox\ is initialized with the pre-trained weights of CountGD. Its MLP box detection heads are then fine-tuned on the FSC-147 training set for 30 epochs with early stopping on the validation set. $\lambda_{loc}$, $\lambda_{GIoU}$, and $\lambda_{cls}$ in the loss are set to 5, 2, and 2 respectively using a grid search on the validation set. For \methodName, in Stage~1, we sample frames at 3 fps. In Stage~2, the window size of the temporal filter, $w$, is set to 3 frames, corresponding to one second. The IoU threshold for matching is set to 0.5. Additional implementation details, including detailed analysis of the inference time and memory consumption for each stage with and without the temporal filter, 
are given in the appendix. 

\subsection{Datasets \& Metrics}\label{sec:datasets_metrics}

\noindent\textbf{Images:} To evaluate the counting and detection accuracy of state-of-the-art detection and counting models in crowded scenes, we use FSCD-147~\cite{10.1007/978-3-031-20044-1_20}, which provides bounding boxes for the validation and test sets of the widely established open-world image object counting dataset FSC-147. Exhaustive bounding boxes are not provided for the training set. Each image is annotated with three visual exemplars. To measure the detection accuracy, we follow \cite{10.1007/978-3-031-20044-1_20} by providing the mean average precision over thresholds 0.5 to 0.95 (AP) and the average precision at the IoU threshold of 0.5 (AP50). We also report the image-based count MAE and RMSE used in \cite{m_Ranjan-etal-CVPR21}. We take the count as the enumeration of the bounding boxes following \cite{Pelhan_2024_NeurIPS}, and report results given text only, three exemplars only, or both, depending on what each method allows. In addition, we report zero-shot counting results for detectors and counters on the ShanghaiTech~\cite{Zhang_2016_CVPR} crowd counting dataset in the appendix. 

\noindent\textbf{Videos:} To evaluate counting accuracy for videos, we report results on \datasetName. Due to the class overlap of the training set of FSC-147 and TAO-Count, we also report results on the subset of videos in TAO-Count with the training classes in FSC-147 removed. To measure counting accuracy for videos, we draw on prior work on object counting in images~\cite{m_Ranjan-etal-CVPR21} that uses the Mean Absolute Error (MAE) and Root Mean Squared Error (RMSE). We define video analogues of these metrics for our new task. More specifically, we define the video MAE and RMSE as: $MAE = (1/N)\sum_{i = 1}^{N}|\hat{y}_{i} - y_{i}|, RMSE = [(1/N)\sum_{i = 1}^{N}(\hat{y}_{i} - y_{i})^2]^{1/2}$, where $N$ is the number of test videos, $\hat{y}_{i}$ is the predicted count for video $X_{i}$, and $y_{i}$ is the ground truth count for video $X_{i}$. In more detail, $y_{i}$ is the number of \emph{unique} objects in the video that match the prompts. Importantly, the MAE and RMSE metrics for counting in videos differ from those used for images. In the video setting, the ground truth count reflects the number of {\em unique object identities}, not the number of detections. This requires both matching and re-identification: objects that reappear across frames must not be double-counted, so repeated detections of the same object must be correctly associated. 

\subsection{Assessing Processing at the Frame Level}
\begin{table}
 
\begin{center}
{\fontsize{9}{11}\selectfont\begin{NiceTabular}{l|c|c|c|c|c} 

    \multicolumn{2}{c}{} & \multicolumn{4}{c}{FSCD-147 Test}\\
   \hline
   \multirow{3}{*}{~~~~~~~Method} & \multirow{3}{*}{Prompt} & \multicolumn{2}{c}{Counting} & \multicolumn{2}{c}{Detection}\\
   & & MAE & RMSE & AP & AP50 \\
   & & $\downarrow$ & $\downarrow$ & $\uparrow$ & $\uparrow$\\
   \hline
  GDINO & text & 54.16 & 157.87 &  11.60 & 17.80 \\
  OWLv2 & text & 41.83 & 149.82 & 22.84 & 35.76\\
  PSeCo & text & 16.58 & 129.77 & \textbf{41.14} & \textbf{69.03} \\
  $\text{DAVE}_{prm}$ & text & 15.52 & \textbf{114.10} & 18.50 & 50.24\\
  CountGD & text & 15.19 & 119.40 & 18.10 & 52.90\\
    CGD-B & text & \textbf{15.01} & 118.16 & 30.44 & 61.56\\
    \hline
    C-DETR & exemp. & 16.79 & 123.56 & 22.66 & 50.57\\
     PSeCo & exemp. & 13.05 & 112.86 & \textbf{43.53} & 74.64 \\
  DAVE & exemp. & 10.45 & 74.51 & 26.81 & 62.82\\
  GeCo & exemp. & \textbf{7.91} & \textbf{54.28} & 43.42 & \textbf{75.06}\\
  CountGD & exemp. & 10.77 & 99.51 & 19.76 & 57.24\\
CGD-B & exemp. & 10.85 & 99.60 & 34.81 & 69.46\\
\hline
CountGD & both & \textbf{10.18} & \textbf{96.20} & 20.50 & 59.40\\
  CGD-B & both & 10.29 & 96.33 & \textbf{36.20} & \textbf{72.39}\\
\hline
\end{NiceTabular}}
\caption{\label{fsc147d_tbl} Results on FSCD-147 for image counting methods that output boxes (for Stage~1). The abbreviations are: \countgdbox\ (CGD-B); C-DETR (Counting-DETR~\cite{10.1007/978-3-031-20044-1_20}). SAM masks are not used here.}
\end{center}
 
\end{table}

\begin{table}
 
\begin{center}
{\fontsize{9}{11}\selectfont\begin{NiceTabular}{c|c|c|c} 

   Temporal Filter & Prompt & MAE $\downarrow$ & RMSE $\downarrow$ \\
   \hline
   \xmark & text & 6.6 & 17.5 \\
   \cmark & text & \textbf{2.6} & \textbf{6.0} \\
\hline
\end{NiceTabular}}
\caption{\label{temporal_filter_ablation} Temporal filter ablation on TAO-Count (Stage~2).}
\end{center}
 
\end{table}

In \cref{fsc147d_tbl}, we evaluate different {\em image} counting and detection methods on FSCD-147 Test given different prompts including text only (`text'), exemplars only (`exemp'), or both together (`both'). Results for FSCD-147 Val are given in the appendix. 
Following~\cite{Pelhan_2024_NeurIPS}, the count is determined by enumerating the bounding boxes, not by summing density maps~\cite{dave} or applying test-time procedures~\cite{countgd} as these operations do not provide boxes. For exemplars, we use the three provided by FSC-147 for each sample. Text descriptions either come from the FSC-147 class names or FSC-147-D~\cite{AminiNaieni23}.  Note, we obtain box predictions from the original CountGD (without the additional training losses added in this paper) by using its full bounding box outputs, rather than, as in the original model, only using the points as output.


From these results, we draw three conclusions: (i) as confirmed by prior work, SoTA detectors like OWLv2~\cite{owlv2} and Grounding DINO~\cite{liu2023grounding} do not work well for the counting setting, where there are many similar and overlapping objects. The caveat here is that these detectors have not been trained on FSC-147, as the counters have. To address this, we also benchmark OWLv2 and Grounding DINO against \countgdbox\ and CLIP-Count~\cite{Jiang2023CLIPCountTT} on the held-out ShanghaiTech counting dataset without any training on ShanghaiTech and show that the counters still significantly beat the detectors in the appendix; 
(ii) Extending CountGD to \countgdbox, significantly improves its detection accuracy while preserving its counting accuracy; (iii) the SoTA model depends on the type of prompt (text/exemplar/both) used. While \countgdbox\ is `a good all-rounder,' it is not the best for all cases. Both \countgdbox\ and PSeCo~\cite{zhizhong2024point} perform competitively in the text-only setting. GeCo~\cite{Pelhan_2024_NeurIPS} is the superior model in the exemplar-only setting, although both \countgdbox\ and DAVE~\cite{dave} are strong contendors. For models that accept both exemplars and text, \countgdbox\ is superior over CountGD for detection. In some cases, the text does add information to the exemplar, by specifying location or color for example (see section 4.5 of \cite{countgd}). However, in settings where this is not the case, like FSC-147 where the text and exemplar both represent the class, GeCo given only exemplars should be used.



\subsection{Assessing processing in the Short Term}

In \cref{temporal_filter_ablation}, we assess the effectiveness of the temporal filter on TAO-Count. Specifically, we report the video-based MAE and RMSE from applying \methodName\ with and without the temporal filter given text only. For the counter, we use \countgdbox, and for the tracker, we use SAM~2.1. The scenes in TAO-Count involve significant motion and blur, inducing false positives. The temporal filter effectively removes these false positives, reducing the MAE and RMSE by over 50\%, improving the counting accuracy significantly.

\subsection{Assessing Processing in the Long Term}

\begin{table}
\begin{center}
{\fontsize{9}{11}\selectfont\begin{NiceTabular}{c|c|c|c|c|c} 
\multicolumn{2}{c}{} & \multicolumn{2}{c}{Penguins} & \multicolumn{2}{c}{Crystals}\\
\hline
\methodName & \multirow{2}{*}{Prompt} & MAE & RMSE & MAE & RMSE \\
Variation &  & $\downarrow$ & $\downarrow$ & $\downarrow$ & $\downarrow$\\
   \hline
  CGD-B/BT & text & 4.3 & 5.5 & 71.6 & 88.1\\
  \textbf{CGD-B/S2} & text &  \textbf{3.3} & \textbf{4.8} & \textbf{69.1} & \textbf{86.0}\\
  CGD-B/S2.1 & text & 4.0 & 5.3 & \textbf{69.1} & \textbf{86.0}\\
  \hline
  \textbf{CGD-B/BT} & exemp & 4.0 & \textbf{4.2} & \textbf{31.1} & \textbf{52.8}\\
    GeCo/S2 & exemp. & 11.7 & 13.1 & 59.6 & 104.2 \\
    GeCo/S2.1 & exemp. & 11.3 & 14.5 & 46.1 & 82.8 \\
  CGD-B/S2 & exemp. &  \textbf{3.3} & 4.8 &  37.4 & 61.0\\
  CGD-B/S2.1 & exemp. & \textbf{3.3} & 4.8 & 33.7 & 59.8\\
    \hline
    CGD-B/BT & both & 7.0 & 8.3 & 12.7 & 16.6\\
  \textbf{CGD-B/S2} & both & \textbf{0.3} & \textbf{0.6} & \textbf{12.0} & \textbf{13.5}\\
  CGD-B/S2.1 & both  & 0.7 & 0.8 & 13.4 & 14.8\\
\hline
\end{NiceTabular}}
\caption{\label{main_results_varied} {\methodName\ video counting results on Science-Count using various prompts, counters, and trackers}. The abbreviations are: \countgdbox\ (CGD-B); ByteTrack (BT);  SAM~2 (S~2); and SAM~2.1 (S~2.1).}
\end{center}
\end{table}

In this section, we
evaluate \methodName's overall video-based counting performance on \datasetName, and compare its performance to strong baselines. For the text descriptions for TAO-Count, we use the category synsets~\cite{tao}. We use the text `human' for MOT20-Count. We use `white crystal' and `penguin' for Science-Count. When exemplars are used, 3--6 exemplars are provided for the first frame of the video and applied to all subsequent frames. The overall results are given in
\cref{main_results_text} and \cref{main_results_varied}.

\noindent\textbf{Baselines:} We compare \methodName\ to strong baselines built from a Multi-Object Tracking (MOT) method and a Video Individual Counting (VIC) method. Specifically, for the first baseline, we repurpose the strong open-world tracker MASA~\cite{masa} implemented with Grounding DINO. The unique tracks are enumerated to estimate the count. For the second baseline, we use the specialized crowd counting method DR.\ VIC~\cite{han2022drvic} trained on HT21~\cite{HT21} and only report results on MOT20-Count, as the other benchmarks in \datasetName\ contain objects other than humans. Note the HT21 training set contains two of the videos in MOT20-Count. 

\methodName\ implemented with \countgdbox\ and 
SAM~2.1
 achieves significantly better counting accuracy than the MASA baseline, as shown in~\cref{main_results_text}. For a fair comparison with MASA, \methodName\ is given text only, even though \methodName\ can also accept exemplars. Remarkably, \methodName\ also significantly improves on the performance of DR.\ VIC, even though DR.\ VIC was trained specifically for counting humans and has been trained on one out of the three videos in MOT20-Count, so it has been trained on one third of the videos we are evaluating it on, while \methodName\ can count other objects in addition to humans and is applied to MOT20-Count zero-shot.
 
 Note that none of the components of \methodName\ (including \countgdbox\ and any hyperparameters) were fine-tuned on the data in \datasetName. We hypothesize the significantly superior performance of \methodName\ can be attributed to (i) leveraging the image-based \emph{counting} model \countgdbox\ rather than the Grounding DINO \emph{detection} model to handle crowded scenes. \countgdbox\ extends Grounding DINO by fine-tuning on counting data and adding modules to enable visual exemplar inputs, meaning it is more accurate and capable than Grounding DINO at counting in video frames; (ii) removing false positives with a temporal filter; and (iii) effectively leveraging video foundation models like SAM~2.1.  
 
 In \cref{main_results_varied}, we compare different variations of \methodName\ implemented with different combinations of counters/trackers and given different prompts. For the counters, we use \countgdbox\ and GeCo, and for the trackers, we use ByteTrack~\cite{bytetrack}, SAM~2 and SAM~2.1. For the ByteTrack variant, we rely on the tracker to detect new objects and do not apply the temporal filter, since there are no segmentation masks. We note the exemplar-only performance is better than text-only, and providing both prompts is the best, showing \methodName\ effectively benefits from more information about the object. We find that while GeCo works well for images, it is not as accurate as \countgdbox\ on videos. SAM~2.1 performs significantly better than SAM~2 in the exemplar-only setting for Crystals. However, it falls slightly behind SAM~2 for the other settings. 
 
 As shown in \cref{fig:qualitative_results}, \methodName\ counts in dense scenes, detects new objects while retaining old ones, and counts deforming objects. Errors can occur due to false negatives from the counting model and tracker re-identification challenges. Scenes with many occlusions and similar instances can cause higher errors due to more of these cases.

\section{Conclusion}
We present the novel task of open-world object counting in videos together with a new model, \methodName, and a new dataset, \datasetName, to test the model. \methodName\ inputs flexible visual exemplar and text prompts and outputs both frame-level counts and a global count indicating the number of unique objects in the video that match the prompts. \methodName\ will continue to benefit from better trackers and class-agnostic detection-based counting models, as they can easily be plugged into the framework we have proposed. 


\section{Acknowledgments}
The authors would like to thank Dr Tom Hart and Penguin Watch for the videos in the Penguins (Science-Count) benchmark, Dr Enzo Liotti for the videos in the Crystals (ScienceCount) benchmark and Shun Yang for the curation and preparation of these videos, Jer Pelhan for his extensive support of GeCo, and Siyuan Li for his extensive support of MASA. This research is funded by an AWS Studentship, the Reuben Foundation, a Qualcomm Innovation Fellowship (mentors: Dr Farhad Zanjani and Dr Davide Abati), the AIMS CDT program at the University of Oxford, EPSRC Programme Grant VisualAI EP/T028572/1, and a Royal Society Research Professorship RSRP$\backslash$R$\backslash$241003.
\small\bibliography{longstrings, vgg_local, other}
\appendix
\renewcommand{\thesection}{Appendix \Alph{section}}
\setcounter{section}{0}
\section*{Appendix}

\noindent A \quad \underline{Additional Quantitative Results} \dotfill\quad\\
\\
 \quad A.1 \quad ShanghaiTech Counting Results \dotfill  \\
\\
 \quad A.2 \quad FSCD-147 Validation Set Results \dotfill  \\
\\
\quad A.3 \quad Temporal Window Length $w$ Sensitivity \dotfill \\
\\
\quad A.4 \quad Frame Rate Sensitivity \dotfill \\
\\
\noindent B \quad \underline{Further Analysis of Baseline Performance} \dotfill\quad\\
\\
 \quad B.1 \quad MASA \dotfill  \\
 \\
 \quad B.2 \quad DR. VIC \dotfill \\
 \\
\noindent C \quad \underline{Further Implementation Details} \dotfill\quad\\
\\
\quad C.1 \quad \methodName\ Implementation Details \dotfill  \\
\\
\quad C.2 \quad Baseline Implementation Details \dotfill  \\
\\
\quad C.3 \quad New Object Detection \dotfill \\
\\
\quad C.4 \quad MAE \& RMSE for Multi-Class Videos \dotfill\\
\\
\noindent D \quad \underline{Computational Requirements} \dotfill\quad\\
\\
\noindent E \quad \underline{Further Dataset Details} \dotfill\quad\\
\\
\quad E.1 \quad TAO-Count \dotfill  \\
\\
\quad E.2 \quad MOT20-Count \dotfill  \\
\\
\quad E.3 \quad Science-Count \dotfill  \\
\\
\noindent F \quad \underline{Clarifications} \dotfill\quad\\
\\
\quad F.1 \quad Open-World Repetition Counting in Videos \dotfill \\
\\
\quad F.2 \quad ``Open-World" vs ``Open-Vocabulary" \dotfill \\
\\
\noindent G \quad \underline{Discussion and Possible Extensions} \dotfill\quad\\
\\
\quad G.1 \quad Boxes vs. Points as Unambiguous Prompts \dotfill\quad \\
\\
\quad G.2 \quad Causality \dotfill\quad\\
\\
\quad G.3 \quad Object Quotas of Counters \& Detectors \dotfill\quad \\
\\
\quad G.4 \quad Limitations \dotfill\quad\\

\setcounter{secnumdepth}{2}

\renewcommand{\thesection}{\Alph{section}}
\renewcommand{\thesubsection}{\thesection.\arabic{subsection}}

\makeatletter
\renewcommand\section{
  \@startsection{section}{1}{\z@}%
  {-2.0ex \@plus -0.5ex \@minus -.2ex}
  {1.0ex \@plus 0.3ex}
  {\normalfont\large\bfseries\raggedright}
}
\makeatother
\section{Additional Quantitative Results}
Here we include additional quantitative results for which there was no room in the main paper.

\subsection{ShanghaiTech Counting Results}

\begin{table}
\begin{center}
{\fontsize{9}{11}\selectfont\begin{NiceTabular}{l|c|c|c|c|c} 

    \multicolumn{2}{c}{} & \multicolumn{4}{c}{ShanghaiTech Test}\\
   \hline
   \multirow{3}{*}{~~~~~~~Method} & \multirow{3}{*}{Prompt} & \multicolumn{2}{c}{Part A} & \multicolumn{2}{c}{Part B}\\
   & & MAE & RMSE & MAE & RMSE \\
   & & $\downarrow$ & $\downarrow$ & $\downarrow$ & $\downarrow$\\
   \hline
  GDINO & text & 394.9 & 537.5 &  58.3 & 99.3 \\
  OWLv2 & text & 420.2 & 553.3 & 81.5 & 126.5\\
  CLIP-Count & text & 192.6 & 308.4 & 45.7 & 77.4\\
  \textbf{CGD-B} & text & \textbf{132.2} & \textbf{253.9} & \textbf{32.2} & \textbf{57.9}\\
\hline
\end{NiceTabular}}
\caption{\label{shanghaitech_tbl} Zero-shot Counting results for detection and counting models on the ShanhaiTech crowd counting dataset. `GDINO' is Grounding DINO and `CGD-B' is \countgdbox.}
\end{center}
\end{table}

In \cref{shanghaitech_tbl}, we evaluate the detection models Grounding DINO~\cite{liu2023grounding} and OWLv2~\cite{owlv2}, and the counting models CLIP-Count~\cite{Jiang2023CLIPCountTT} and \countgdbox, on both Part A and Part B of the ShanghaiTech crowd counting dataset~\cite{Zhang_2016_CVPR}. This table clearly shows that the counting models achieve lower counting errors than the detection models without training on any of the images in ShanghaiTech. For the CLIP-Count result, we use the published results from the CLIP-Count paper from training CLIP-Count on FSC-147 and testing it zero-shot on ShanghaiTech. For \countgdbox\ and Grounding DINO, we use the text `human'. For OWLv2, we use the text `person', since it significantly improves its results compared to using `human'.

\subsection{FSCD-147 Validation Set Results}

\begin{table}
\begin{center}
{\fontsize{9}{11}\selectfont\begin{NiceTabular}{l|c|c|c|c|c} 

    \multicolumn{2}{c}{} & \multicolumn{4}{c}{FSCD-147 Val}\\
   \hline
   \multirow{3}{*}{~~~~~~~Method} & \multirow{3}{*}{Prompt} & \multicolumn{2}{c}{Counting} & \multicolumn{2}{c}{Detection}\\
   & & MAE & RMSE & AP & AP50 \\
   & & $\downarrow$ & $\downarrow$ & $\uparrow$ & $\uparrow$\\
   \hline
  GDINO & text & 54.45 & 137.12 &  6.66 & 10.26 \\
  OWLv2 & text & 49.02 & 131.41 & 11.39 & 20.33\\
  PSeCo & text & 23.90 & 100.30 & \xmark & \xmark \\
  $\text{DAVE}_{prm}$ & text & 14.87 & \textbf{59.60} & 16.31 & 47.12\\
  CountGD & text & 12.46 & 67.52 & 13.36 & 43.90\\
    CGD-B & text & \textbf{12.24} & 66.24 & \textbf{27.02} & \textbf{60.73}\\
    \hline
    C-DETR & exemp. & 20.38 & 82.45 & 17.27 & 41.90\\
     PSeCo & exemp. & 15.31 & 58.34 & 32.12 & 60.02 \\
  DAVE & exemp. & 9.75 & \textbf{40.30} & 24.20 & 61.08\\
  GeCo & exemp. & 9.52 & 43.00 & \textbf{33.51} & 62.51\\
  CountGD & exemp. & 9.34 & 51.74 & 15.16 & 47.96\\
CGD-B & exemp. & \textbf{9.27} & 52.15 & 32.09 & \textbf{68.82}\\
\hline
CountGD & both & \textbf{8.69} & \textbf{43.89} & 14.74 & 47.95\\
  CGD-B & both & 8.76 & 44.24 & \textbf{33.10} & \textbf{71.38}\\
\hline
\end{NiceTabular}}
\caption{\label{fsc147d_val_tbl} Results on FSCD-147 Val for image counting methods that output boxes (for Stage~1). \xmark\ indicates results are not provided by the authors. `GDINO' is Grounding DINO, `CGD-B' is \countgdbox, and `C-DETR' is Counting-DETR~\cite{10.1007/978-3-031-20044-1_20}. SAM masks are not used in evaluation. `exemp.' means exemplars only, and `both' means both exemplars and text as prompts.}
\end{center}
\end{table}

In \cref{fsc147d_val_tbl}, we include the results from evaluating object detection and counting models on the validation set of FSCD-147~\cite{10.1007/978-3-031-20044-1_20, m_Ranjan-etal-CVPR21}. For text only, \countgdbox\ achieves the best performance on the validation set. For exemplars only, \countgdbox\ is competitive with GeCo~\cite{Pelhan_2024_NeurIPS}. For both exemplars and text, \countgdbox\ preserves the counting accuracy of CountGD~\cite{countgd} while significantly improving its detection accuracy. As stated in the main paper and as shown in \cref{shanghaitech_tbl}, the detectors perform poorly compared to the counters.

\subsection{Temporal Window Length $w$ Sensitivity}

\begin{table}
\begin{center}
{\fontsize{9}{11}\selectfont\begin{NiceTabular}{l|c|c|c} 
\multicolumn{2}{c}{} & \multicolumn{2}{c}{TAO-Count} \\
\multicolumn{2}{c}{}& \multicolumn{2}{c}{\emph{(10-video subset)}}\\
\hline
Temporal Filter & \multirow{2}{*}{Prompt} & MAE & RMSE  \\
Window Length $w$ &  & $\downarrow$ & $\downarrow$ \\
   \hline
   6 frames& text & 4.4 & 10.3\\
  3  frames& text  & 6.5 & 17.3 \\
     1  frame & text & 14.2 & 34.4 \\
\hline
\end{NiceTabular}}
\caption{\label{temp_filter_window_lngth} \methodName\ video counting results on the first 10 videos in TAO-Count using different temporal filter window lengths and given only text.}
\end{center}
\end{table}

Here we test how the window length $w$ of the temporal filter affects counting accuracy. For this experiment we use the first 10 videos from TAO-Count corresponding to 18 category-video pairs. Due to computational limitations, we do not run this test on the full TAO-Count dataset. The frame rate is held at 3 fps as it was set in the main paper while the window length for the temporal filter is varied from 1-6 frames. In this experiment, \methodName\ is implemented with \countgdbox\ and SAM~2.1.

As shown in \cref{temp_filter_window_lngth}, wider window lengths result in lower counting errors. This could be because in TAO-Count real objects tend to persist for at least two seconds and any frame-level detections that do not persist for this long are false positives. False positives that persist for more than one second (3 frames) but less than two seconds (6 frames) will only be removed by a temporal filter with a window length greater than 3. However, larger temporal filter windows result in lengthier inference times. Therefore, one needs to consider a balance between efficiency and accuracy when determining the length of the temporal filter.

\subsection{Frame Rate Sensitivity}

\begin{table}
\begin{center}
{\fontsize{9}{11}\selectfont\begin{NiceTabular}{l|c|c|c|c|c} 
\multicolumn{2}{c}{} & \multicolumn{2}{c}{Penguins} & \multicolumn{2}{c}{Crystals}\\
\hline
\multirow{2}{*}{Frame Rate} & \multirow{2}{*}{Prompt} & MAE & RMSE & MAE & RMSE \\
 &  & $\downarrow$ & $\downarrow$ & $\downarrow$ & $\downarrow$\\
   \hline
   6 fps & text & 4.0 & 5.4 & 69.1 & 86.0\\
  3 fps & text  & 4.0 & 5.3 & 65.9 & 84.1\\
     1 fps & text & 0.7 & 0.8 & 69.6 & 86.1 \\
\hline
\end{NiceTabular}}
\caption{\label{frame_rate} \methodName\ video counting results on Science-Count using different frame rates and given only text.}
\end{center}
\end{table}

In \cref{frame_rate}, we report the counting errors of \methodName, implemented using \countgdbox\ and SAM~2.1, under different frame sampling rates on the Science-Count dataset. In this experiment, we provide text-only prompts to \methodName\ (i.e., no exemplars are used). The original setup in the main paper uses a frame rate of 3~fps with a temporal window length $w$ of 3 frames (i.e., 1 second) for Penguins, and 6~fps with a $w$ of 3 frames (0.5 seconds) for Crystals. Here, we maintain a fixed \textit{temporal duration} for the filter window---1 second for Penguins and 0.5 seconds for Crystals---across all tested frame rates.

For Penguins, we observe a significant improvement in counting accuracy as the frame rate decreases. Sampling at 6~fps leads to substantially higher errors than at 1~fps. A likely explanation is that higher frame rates introduce more noisy frame-level predictions (due to small motion or detection artifacts), which propagate and accumulate throughout the full video. Sampling fewer, more spaced-out frames suppresses this noise, making the aggregated prediction more robust.

For Crystals, the relationship between frame rate and counting accuracy is less straightforward. Reducing the frame rate from 6~fps to 3~fps improves performance, but further reducing it to 1~fps slightly increases the error. One explanation is that the 1~fps setting may cause the model to miss crystals that appear and disappear within a 1-second window, increasing counting error. Alternatively, lower temporal resolution could hurt tracking and association by skipping over intermediate frames.

Overall, these results indicate that the impact of frame rate is \textit{dataset-specific}. Stable scenes like Penguins benefit from reduced frame rates, while rapidly changing scenes like Crystals require a balance between temporal coverage and noise suppression. Selecting the frame rate based on a dataset-specific validation set to maximize both accuracy and computational efficiency is a promising direction for future work.

\section{Further Analysis of Baseline Performance}
\subsection{MASA~\cite{masa}}
For MASA, we notice a bias towards predicting higher object counts than the ground truth on MOT20-Count. Grounding DINO~\cite{liu2023grounding} produces unstable detections on the frames in MOT20-Count, since it is poor at detecting objects in crowded scenes. This results in lost tracks and the initialization of new tracks with new IDs, erroneously increasing the count. \methodName\ mitigates this issue by leveraging counting models that provide more persistent and stable detections in crowded scenes than detection models do. \methodName\ also leverages the more complex memory and occlusion prediction mechanisms of SAM~2.1~\cite{ravi2024sam2} to retain and re-identify objects.

\subsection{DR.\ VIC~\cite{han2022drvic}}
DR.\ VIC is trained specifically for crowd counting and uses one of the MOT20-Count videos for validation and one for training, leaving only one-third of the videos in MOT20-Count as truly held out. Despite this, we test DR.\ VIC on all three videos in MOT20-Count, giving it an unfair advantage over \methodName, which is tested zero-shot on MOT20-Count. Then why does \methodName\ perform so much better than DR.\ VIC on MOT20-Count? We hypothesize this is because \methodName\ effectively leverages foundation models such as CountGD~\cite{countgd} (that uses the frozen image and text backbones of Grounding DINO, while finetuning the remaining modules on counting data) and SAM~2.1, which are trained on large quantities of data. On the other hand, DR.\ VIC is only trained and validated on the four training videos in HT21~\cite{HT21}. This results in \methodName\ being a more accurate and general counting model that achieves superior performance on MOT20-Count than DR.\ VIC despite DR.\ VIC's unfair advantage.

\section{Further Implementation Details}

\subsection{\methodName\ Implementation Details}
Here we detail the implementation details for each of the variants of \methodName\ used in table 5 of the main paper.

\paragraph{\countgdbox/SAM~2.1 (Main) Variant.} For the main variant of \methodName\ used for comparing against baselines we set the hyperparameters as follows. For \countgdbox, we set the confidence threshold to 0.23, which is the same as the threshold for CountGD~\cite{countgd}. We do not do any tuning of this threshold. For training, we start with the pretrained CountGD checkpoint published by the authors and use the same open-source training code, number of epochs (30) and seed. We set $\lambda_{loc}$, $\lambda_{GIoU}$, and $\lambda_{cls}$ to 5, 2, 2 respectively. These settings were selected after trying the combinations $(\lambda_{loc}=5, \lambda_{GIoU}=2, \lambda_{cls}=2)$, $(\lambda_{loc}=1, \lambda_{GIoU}=2, \lambda_{cls}=5)$ and $(\lambda_{loc}=1, \lambda_{GIoU}=0, \lambda_{cls}=5)$ and choosing the combination that achieved the lowest counting MAE on the FSC-147~\cite{m_Ranjan-etal-CVPR21} image counting validation set. We only train the model once and then use it for inference on \datasetName\ without any fine-tuning on the videos in \datasetName. For SAM~2.1~\cite{ravi2024sam2}, we use the $sam2.1\_hiera\_large.pt$ checkpoint and code published by the authors. 

To speed up the experiments, we set the frame sampling rate for all the videos in \datasetName\ to three frames per second, except the videos in the Crystals dataset. For those videos, the sampling rate is 6-7 frames per second, due to how rapidly the crystals form and disappear. The videos were provided to us pre-sampled at this frame rate. The window size $w$ for the temporal filter is set to 3 for all the videos in \datasetName, corresponding to one second for most of the videos. These hyperparameters were not tuned for optimal accuracy on \datasetName. Future work could include developing a video counting validation set for this purpose.

\paragraph{\countgdbox/SAM\,2 Variant.} This variant is the same as the main variant, but uses the published $sam2\_hiera\_large.pt$ checkpoint instead of the $sam2.1\_hiera\_large.pt$ checkpoint published by the authors of SAM~2.

\paragraph{GeCo/SAM~2.1 Variant.} This variant is the same as the main variant except that we use the GeCo~\cite{Pelhan_2024_NeurIPS} counting and detection model. For GeCo, we use the pretrained checkpoint and code published by the authors. We use the same division threshold of 8 as the authors for almost all of the videos in \datasetName\ except for Crystals due to too many false positives. For Crystals we set the division factor to 2. 

\paragraph{GeCo/SAM\,2 Variant.} This variant is the same as the GeCo/SAM~2.1 variant except we use the published $sam2\_hiera\_large.pt$ checkpoint instead of the $sam2.1\_hiera\_large.pt$ checkpoint published by the authors of SAM~2.

\paragraph{\countgdbox/ByteTrack Variant.} This variant is the same as the main variant except we replace the tracker with ByteTrack~\cite{bytetrack}. Since ByteTrack does not provide segmentation masks, we do not use the original new object detection logic and temporal filter from \methodName. Instead, we use ByteTrack's inbuilt tracking mechanisms. We still use \countgdbox\ to produce detections, which are then tracked with ByteTrack. The number of unique tracks identified in the video are enumerated to predict the final count. We use the implementation of ByteTrack provided by the Roboflow Supervision library.

\subsection{Baseline Implementation Details}
\paragraph{MASA~\cite{masa}.}
For the MASA baseline, we used the Grounding DINO~\cite{liu2023grounding} checkpoint and code published by the authors with the Grounding DINO features and detections. Before deploying MASA on \datasetName, we reproduced the results reported in the MASA repository for the pretrained checkpoint. We then adapted the code for the published demo to test MASA on \datasetName. In our adapted code, the unique tracks output by MASA are enumerated to obtain the final count given the same text as all the methods compared to it. We set the $memo\_tracklet\_frames$ to 300 to help with retaining object identities for objects that disappear for long periods. All the other hyperparameters are set to the default values in the published demo. We use these hyperparameters for all the videos in \datasetName. 

\paragraph{DR. VIC~\cite{han2022drvic}.}
For DR.\ VIC, we use the code and HT21 checkpoint published by the authors. Before deploying DR.\ VIC on MOT20-Count, we reproduce the published results on the HT21 test set. As noted above, the authors train on the MOT20-05 sequence in MOT20-Count (HT21-02 in HT21), in addition to the other training videos, and use the MOT20-01 sequence in MOT20-Count (HT21-01 in HT21) for validation. 
Here, we document and discuss our results using DR.\ VIC for each video of MOT20-Count for reference.

\begin{table}[h!]
\begin{center}

\begin{NiceTabular}{l|c|c|c|c} 
\multirow{2}{*}{Sequence} & \multirow{2}{*}{Pred} & \multirow{2}{*}{GT} & Absolute & Relative\\
& & & Error $\downarrow$ & Error $\downarrow$\\
\hline
MOT20-01$^{V}$ & 112.8 & 80 & 32.8 & 0.4\\
MOT20-02 & 418.1 & 278 & 140.4 & 0.5\\
MOT20-05$^{T}$ & 1329.7 & 1203 & 126.7 & 0.1\\
\hline
\end{NiceTabular}
\caption{ \label{dr_vic_mot20} DR.\ VIC counting results on the sequences in MOT20-Count. The $V$ superscript on MOT20-01 indicates it was used for validation, and the $T$ superscript on MOT20-05 indicates it is in the training set for DR.\ VIC.}
\end{center}
\end{table}

As expected, in \cref{dr_vic_mot20}, it shows that DR.\ VIC achieves the highest error on the held-out MOT20-02 sequence. In fact, this error is very close to the reported mean absolute error of 141.1 (table 2 of ~\cite{han2022drvic}) of DR.\ VIC on HT21. 
The predicted count of 112.8 is also almost exactly the reported count of 113.0 on the equivalent HT21-01 sequence in the same table 2 of the DR.\ VIC paper. Finally, the authors include a qualitative result of DR. VIC's performance on a clip of MOT20-05 in the published repository, and the result for MOT20-05 in \cref{dr_vic_mot20} is consistent with this result.

\subsection{New Object Detection}

\begin{figure}[h]
  \centering
  \includegraphics[width=\linewidth]{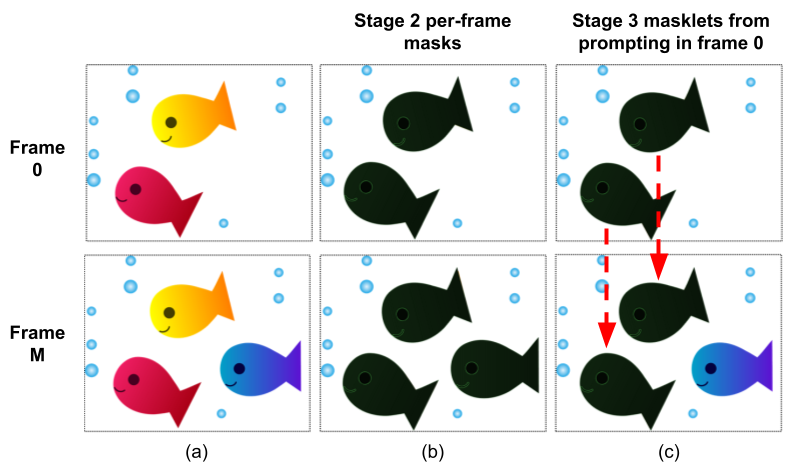}
   \caption{New object detection logic.  A new blue fish that did not appear in frame 0 is detected in frame M, since no propagated mask from frame 0 overlaps with the blue fish in frame M (c).}

   \label{fig:new_obj}
\end{figure}

In \cref{fig:new_obj}, we illustrate how \methodName\ is able to detect new objects. In more detail, in frame 0 \cref{fig:new_obj} (a) there are initially two fish. In frame M, a new blue fish appears. In \cref{fig:new_obj} (b) masks are predicted and verified for the fish independently for each frame in Stage~2, correctly segmenting the fish in each frame. In \cref{fig:new_obj} (c), the two masks for the fish in frame 0 are propagated in Stage~3 from frame 0 to frame M (indicated by the red arrows), and the masklets overlap with the independent object masks for the two fish. However, there is no mask propagated to overlap with the blue fish, since it was not in frame 0. Hence, the blue fish is considered a new object.

\subsection{MAE \& RMSE for Multi-Class Videos}
While in the main paper, we define the video-based MAE and RMSE for the case where only one class is labeled per video, here we explain how these metrics are calculated when multiple classes are labeled per video. Instead of averaging over all videos, in the multi-class case, we average over all category-video pairs. This means that if a single video has multiple objects, then the error from each category-video pair is treated equally. For example, if video one has cats and dogs labeled and video two has only fish labeled, then the total MAE and RMSE for these two videos would consist of three terms, each weighted equally. The first term would correspond to the error from video one for counting cats. The second term would come from video one for counting dogs. The third term would come from video two for counting fish. Each term would have a weight of one. Notice that this also means that because video one has more objects labeled, errors from it contribute more to the final MAE and RMSE than errors from video two. 

\section{Computational Requirements}
\methodName\ runs inference in three stages. Here we detail the time and memory requirements of \methodName\ for each stage. We run \methodName\ implemented with \countgdbox\ and SAM~2.1 (large) on 1 NVIDIA RTX 3090 GPU with 24 Gb of GPU memory and 16 CPU cores. We use the Ubuntu 22.04.4 LTS operating system. We use the latest versions of the CountGD and SAM~2 codebases. The time for each stage given these resources for a 30-second video of 39 penguins is detailed in column 1 of \cref{compute_table}. The projected time for each stage given 8 GPUs and 128 CPU cores is given in column 2 of \cref{compute_table}. In general, the time and memory requirements of \methodName\ are governed by the requirements of the base counting and tracking models.

\begin{table*}[t!]
\begin{center}
\caption{ \label{compute_table} Inference time for running \methodName\ on a 30-second video of 39 penguins. Stage~2a refers to tracking forwards and backwards in time for each detection in Stage~1. Stage~2b refers to matching the propagated masks from Stage~2a with the per-frame masks from Stage~1 to identify and remove false positives. All projected timings in the rightmost column are estimated analytically; only the 1-GPU, 16-CPU results are implemented and measured directly.}

\begin{NiceTabular}{c|c|c} 
\hline
& 1 GPU, 16 CPU Cores & 8 GPUs, 128 CPU Cores (Projected Analytical Estimate)\\
\hline
Stage~1 & 1m 19s & 10s\\
Stage~2a & 11m 19s & 1m 25s\\
Stage~2b & 7m 31s & 56s\\
Stage~3 & 6m 37s & 2m 27s\\
\hline
Total + Stage~2 & 26m 46s & 4m 58s\\
Total - Stage~2 & 7m 56s & 2m 37s\\
\hline
\end{NiceTabular}
\end{center}
\end{table*}

\paragraph{Stage~1.} 
In Stage~1, the counting and segmentation models are applied to each frame independently. Therefore, operations can be completed in parallel over multiple batches and GPUs. For a fixed batch size, the time and memory for this stage scales linearly with the number of frames (video length) and the number of objects. In our setup, as shown in column 1 of \cref{compute_table}, Stage~1 takes 1 minute and 19 seconds and 17 GB of GPU memory for a 30-second video of 39 penguins. Given 8 GPUs, this time is divided by 8 in column 2. 

\paragraph{Stage 2.} 
In Stage~2, a temporal filter is applied to each detection in Stage~1 to remove false positives. We break up Stage~2 into Stage~2a, which uses GPU(s), and Stage~2b, which uses only CPU(s).

In Stage~2a, each detection in Stage~1 is tracked forward $w$ frames and backward $w$ frames from the original frame the detection is in. This results in processing $2w - 1$ unique frames for each detection. There is no redundant processing in that for each detection, we track it in a particular frame only once. Since each detection is tracked independently and frames are processed independently, operations can be completed in parallel over multiple batches of frames and objects over multiple GPUs. For a fixed batch size, the time for this stage scales linearly with the number of frames, the number of objects, and the filter width $w$. The memory scales linearly with the number of objects. In our setup, as shown in column 1 of \cref{compute_table}, Stage~2a takes 11 minutes and 19 seconds and 17 GB of GPU memory for a 30-second video of 39 penguins. Given 8 GPUs, this time is divided by 8 in column 2.  

In Stage~2b, each propagated mask from Stage~2a is matched with per-frame masks from Stage~1. Since masks are checked independently, they can be processed in parallel. This stage requires no GPUs. The time for this stage scales linearly with the number of frames and the number of objects. In our setup, as shown in column 1 of \cref{compute_table}, Stage~2b takes 7 minutes and 31 seconds for a 30-second video of 39 penguins. Given 128 CPUs, this time is divided by 8 in column 2.  

We also consider the inference time for \methodName\ when Stage~2 is removed in the last row of \cref{compute_table}. In applications where processing time is more important than removing false positives, \methodName\ can be applied without Stage~2, resulting in faster execution times.

\paragraph{Stage 3.}
In Stage~3, the video is processed iteratively, so there is limited opportunity for parallelism. Furthermore, the degree of parallelization possible depends on the tracking model. SAM~2 and SAM~2.1 treat each object independently. This means that masks from one object cannot influence the masks from another object. As a result, the objects can be propagated in parallel, in separate batches, and on separate GPUs. 

In a non-causal setup, all the objects detected in the first frame can be propagated in parallel for the whole video. Then, for each frame after the first frame, \methodName\ checks for new objects, propagates them in parallel, and combines them with the masks of the other objects. This continues iteratively throughout the video. Since the frames are processed iteratively, and each object is propagated once, the time and memory for this phase scale linearly with the number of frames and the number of objects.

In our setup, as shown in column 1 of \cref{compute_table}, Stage~3 takes 6 minutes and 37 seconds and 17 GB of GPU memory for a 30-second video of 39 penguins. To project the time required given 8 GPUs, we divide time required to propagate the objects in the initial frame by 8 and add that to the time required to process the rest of the video in our setup with 1 GPU.


\section{Further Dataset Details}
\begin{figure*}[ht!] 
  \centering
  \includegraphics[width=\linewidth]{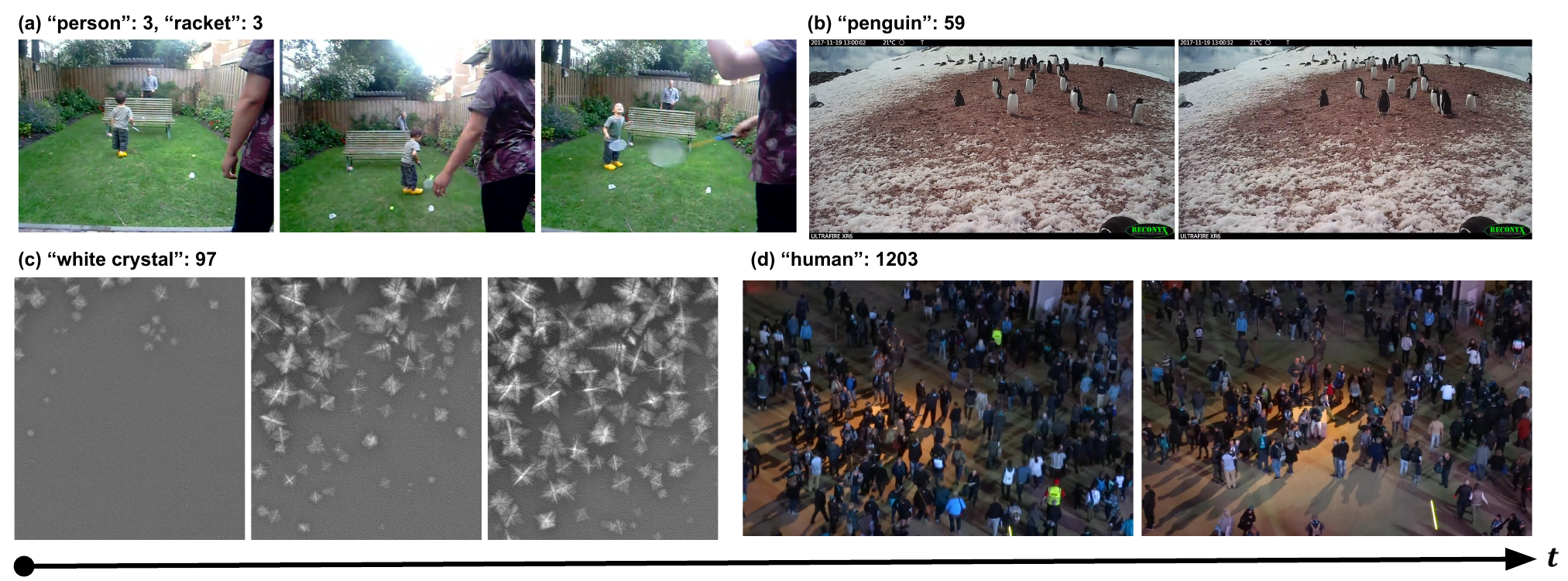}
  \caption{Examples from \datasetName. (a) TAO-Count includes multiple moving objects (rackets and players). Science-Count contains videos of (b) similar-looking penguins of varying sizes and (c) crystals in x-ray videos that change structure over time. (d) MOT20-Count involves counting over a thousand people in dense crowds. \datasetName\ poses diverse, high-density, and dynamic challenges for open-world counting.}
  \label{fig:datasets_examples}
\end{figure*}

\begin{table*}[h!]
\begin{center}

\fontsize{9}{11}\selectfont\begin{NiceTabular}{l|c|c|c|c|c} 
\hline
   Dataset & Video? & \# of Samples & \# of Classes & Min. \# Objects per Sample & Max. \# of Objects per Sample \\
   \hline
   FSC-147 & \xmark & 6135 & 147 & 7 & 3731 \\
   CARPK & \xmark & 1448 & 1 & 1 & 188 \\
   VisDrone & \cmark & 263 & 10 & - & - \\
   Mall & \cmark & 1 & 1 & 60,000 & 60,000  \\
   \datasetName & \cmark & 370 & 141 & 1 & 1203 \\
\hline
\end{NiceTabular}
\caption{ \label{comparison_datasets_tbl} Comparing \datasetName\ to the other counting datasets: FSC-147, CARPK~\cite{Hsieh2017DroneBasedOC}, VisDrone~\cite{visdrone}, and Mall~\cite{mall_dataset}. A `sample' refers to a single image if considering an image counting dataset or a single video if considering a video counting dataset. `-' indicates that information is not accessible. Object counting datasets with large numbers of categories, as is expected in the open-world setting, are not available for our new task. While FSC-147~\cite{m_Ranjan-etal-CVPR21} covers 147 categories, it does not support videos. While VisDrone~\cite{visdrone} and Mall~\cite{mall_dataset} support videos, they are limited to 1-10 categories. Our proposed \datasetName\ overcomes this limitation by covering a large number of categories while also supporting videos.}
\end{center}
\end{table*}

\begin{figure}[ht!]
  \centering
  \includegraphics[width=\linewidth]{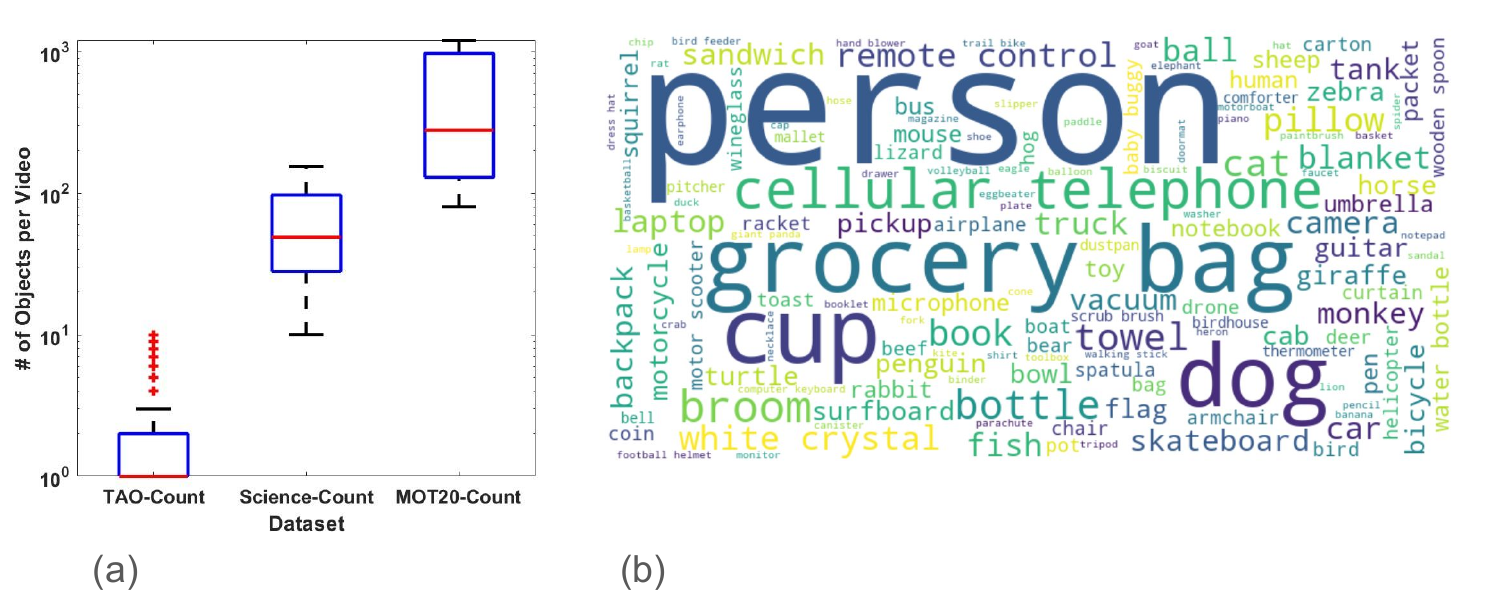}
  \caption{In (a), box plots show that the \datasetName\ covers a large range of counts from 1-1203 objects per video. In (b), a word cloud shows the diversity of object classes covered. The `person' class is quite common because it is one of the most popular exhaustively annotated classes in TAO-Count.}
  \label{fig:data_stats}
\end{figure}

Here we include further details on \datasetName. In \cref{fig:datasets_examples}, we include sample frames from \datasetName. In \cref{comparison_datasets_tbl}, we compare \datasetName\ to other counting datasets. In \cref{fig:data_stats}, we show \datasetName\ spans a wide range of categories and counts. While \datasetName\ includes more videos with fewer than ten objects than those with more, this imbalance reflects a common challenge in object counting datasets. For example, FSC-147~\cite{m_Ranjan-etal-CVPR21}, a well-established image counting dataset, also exhibits significant imbalance due to the inherent difficulty of annotating high object counts. This challenge is further amplified in videos, where annotating counts across hundreds of frames involves considerable effort. Despite this, we include videos with annotations for up to 1203 objects, a large number for video counting benchmarks. To avoid any bias toward sparse scenes, we report results separately for TAO-Count (1-10 objects) and MOT20-Count (80-1203 objects). 

\subsection{TAO-Count}\label{tao_process}
TAO~\cite{tao} contains 2,907 videos with tracks labeled for 833 different object categories. Due to its diversity of object types, we build TAO-Count on top of TAO to test how well open-world object counting methods generalize to many different categories. However, TAO has certain limitations. Firstly, it has on average 5.9 tracks per video, and the maximum number of tracks per video in our filtered subset of TAO is 10, which is considered very low for counting. For example, FSC-147~\cite{m_Ranjan-etal-CVPR21}, the standard open-world object counting dataset for images, has on average 56 objects and at most 3,731 objects per image. Additionally, not all objects in TAO are exhaustively annotated, which is necessary for counting. To address the former issue, we also propose MOT20-Count. To solve the latter issue, we filter out categories and videos that are not exhaustively annotated from the TAO validation set. This process results in 357 videos covering 139 categories in TAO-Count. We do not include counts for the AVA and HACS videos as these videos require third-party approvals to obtain. To obtain the text input for the categories, we use the category synsets. 

\subsection{MOT20-Count}\label{mot20_process}
MOT20~\cite{mot20} contains 8 videos with heavily crowded scenes ranging from 17 seconds to a little over 2 minutes. To build MOT20-Count, we use three of the training videos, each with 80 to 1203 humans for evaluating counting methods on crowded scenes. The complete tracking annotations for MOT20~Test, which are required to get the final counts, are not publicly available as they are for MOT20~Train. This restriction is in place to prevent test set leakage and preserve the integrity of MOT20 as a well-established tracking benchmark. Despite this limitation, we chose to evaluate on a subset of the MOT20 training videos because they include dense scenes with many occlusions and overlapping pedestrians, critical for evaluating object counting in challenging conditions. To maintain fairness, we do not train any of the components of \methodName\ on MOT20. 

The original MOT20 ground truth annotations include the ‘crowd’ class, where regions of the video considered too densely populated are grouped together into a single bounding box. Furthermore, static people, people in vehicles, and children in strollers are omitted from the MOT20 evaluation procedure, even though their ground truth boxes are provided. For the counting task, given the text ‘human,’ the model should count \emph{all} the humans. Therefore, in MOT20-Count, we omit videos from MOT20 that include instances of the ‘crowd’ class. For the remaining videos, we add the numbers of unique trajectories for the ‘pedestrian’ class, ‘static person class,’ and ‘person on vehicle’ class to the count. We examine the bounding boxes for the remaining classes and add any humans that do not fall into the previous categories (e.g., children in strollers and disabled people in electric wheelchairs) to the count. We do not include children in strollers that are not visible throughout the entire video. The result of this process is what we consider to be the ground truth count.

Another note is that the original MOT20 annotations do not take into account re-identification--if humans leave and reenter the scene, they are relabeled as new instances. However, the maintainers of MOT20 have ensured us that such cases should happen very rarely if at all in the dataset. Furthermore, we incorporate re-identification challenges into TAO-Count. For example, there are videos where humans leave for more than 2 seconds and return to the scene with the same identity label in TAO-Count. Therefore, \datasetName\ does explicitly challenge counting methods with re-identification tasks. 

Finally, we compare our MOT20-Count to an existing crowd counting dataset that also has a comparably high average human count ($> 500$ per video) to show why it was necessary to build our own dataset. The tracking dataset CroHD (also known as HT21)~\cite{HT21} has been used as a Video Individual Counting (VIC) dataset. However, the ground truth counts used from this dataset by VIC methods like DR. VIC~\cite{han2022drvic} have issues. Looking at table 2 of \cite{han2022drvic}, it is clear that the {\em total number of tracks} rather than the {\em total number of humans} is being used as the ground truth count. These tracks include tracks for statues and human faces on clothing. On the other hand, in MOT20-Count, we only count the number of living humans. At the same time, we leverage existing annotations from MOT20, to avoid reinventing existing datasets. Furthermore, these datasets do not include significant re-identification challenges, which is one of the reasons that it was necessary to add TAO-Count to our suite of benchmarks as it contains significant re-identification challenges. Further comparisons of our full \datasetName\ to other benchmarks are presented in \cref{comparison_datasets_tbl}.

\subsection{Science-Count}\label{science_count_process}
\textbf{Penguins.} To test \methodName\ on real-world biodiversity applications, we include three 10-30 second videos of penguins in their natural habitats. These videos were initially taken to monitor the penguin population in a particular region over time. We sample the videos at 3 frames per second and annotate each frame using the VGG Image Annotator~\cite{Dutta19a}. Specifically, we add a single point inside each penguin for every frame. Penguins are then matched manually across frames. The total number of unique penguins identified in each video is the ground truth count for that video. Each video has 28-59 penguins.

\noindent \textbf{Crystals.} In Science-Count, we include new videos from the very challenging real-world setting of counting crystals that form from liquid metal alloys. Because this process is very rapid, we do not downsample the frames for annotation. Instead, we annotate the frames at 6-7 frames per second. The bounding boxes of the x-ray images are already provided. However, as the crystals form, many of them disappear while new ones enter. To account for this, we calculate the ground truth count by adding the number of crystals that disappear to the number of crystals in the final frame. In this setting, no crystals re-enter the scene, as they all move in the same direction out of the camera's field of view and never return. We provide data for seven videos each with 10-154 crystals.

\section{Clarifications}

\subsection{Open-World Repetition Counting in Videos}
While open-world {\em repetition} counting in videos is an interesting problem with existing datasets like OVR~\cite{ovr} for testing methods on it, our work is distinct from this task. 
Methods for \emph{repetition} counting~\cite{counting_out_time, live_rep_count, sinha2024every, zhang2021repetitive}, as opposed to \emph{object} counting, count the number of times a recurring action or motion event (e.g., a single jumping jack, clap of a hand, bounce of a ball) occurs in a video. Recent work~\cite{sinha2024every} has followed the evolution of models for counting objects in images, by developing models where an exemplar clip can specify the type of repetition to count. However, unlike objects, events are defined by temporal, rather than spatial, boundaries, and the essence of the repetition counting task is {\em temporal self-similarity}. Consequently, the repetition counting task has more in common with object counting in images (which is concerned with
spatial self-similarities) than with object counting in videos.

\subsection{``Open-World" vs ``Open-Vocabulary"}
Here we make an important note about terminology. We notice that in the object counting literature~\cite{AminiNaieni23, countgd, Liu22}, {\em open-world counting} refers to the task of counting instances of an object class specified at test time via textual or visual prompts. These counting models are {\em open-world} because they generalize beyond a fixed vocabulary, as the object of interest may belong to a class not seen during training. However, crucially, the category is still explicitly provided as input at inference time, either as text or exemplar. We adopt this definition of {\em open-world} in our work, since \methodName\ builds on counting literature.

However, this usage of {\em open-world} differs from that in some earlier literature. The origins of the formal definition of `open world' are from \cite{owr}. This paper defines an open-world recognition system as one that recognizes instances of both \emph{known} and \emph{unknown} classes, marks the unknown objects as `unknown,' obtains class labels for these unknown objects, and incrementally learns from these instances such that they become `known.' Similarly, early works in open-world object detection \cite{joseph2021open} aim to discover and learn novel object categories without labels at first, marking them as `unknown,' and incrementally learning them later. However, more recent works in detection~\cite{owlv2, owl} use the term `open-world' to describe a detection model that can detect objects unseen during training by accepting textual prompts (i.e., labels) describing the object. Importantly, this means that the labels of the unknown objects must be provided for the model to detect them, which differs from the original formal definition of `open world' in \cite{owr}. The OWL paper~\cite{owl} uses the terms `open-world' and `open-vocabulary interchangeably to describe this prompt-based setting as we do. In fact `OWL' stands for `Open-World Localization.' Like OWL, in our case, the category is always specified by the user, and the challenge lies in handling domain shifts, visual diversity, and lack of training-time exposure to the category.

\section{Discussion and Possible Extensions}

\subsection{Boxes vs.\ Points as Unambiguous Prompts}
In general, bounding boxes are less ambiguous specifications of the object to count than points are. In fact, the whole field of visual-exemplar-based counting relies on this idea~\cite{Gong2022ClassAgnosticOC, Liu22, Lu18, 10.1007/978-3-031-20044-1_20, m_Ranjan-etal-CVPR21, Shi2022RepresentCA, yangClassagnosticFewshotObject2021,  You_2023_WACV, low_shot, Lin_2022_BMVC}. For these methods, it is assumed that a `visual exemplar', bounding box over one example instance, is an unambiguous specification of the object to count. However, it is interesting to note that in some (very rare) cases, overlapping objects may share the same tight bounding box. For example, if one person stands directly behind another person with little to no part of their body showing, a bounding box could refer to either (a) the person in the front, (b) the person in the back, or (c) both people simultaneously (e.g., counting \emph{pairs} of people rather than each person, one at a time). However, usually the IoU between bounding boxes of different objects will be less than one, allowing \methodName\ to unambiguously identify each object separately. This is much better than points, where the object can be identified as its whole or subparts.  Future work could consider using an object {\em segmentation} to specify the prompt.

\subsection{Causality}
Here we discuss the aspect of causality and what would be required to make \methodName\ causal. By `causal' we mean that \methodName\ provides the global count -- the number of unique objects in the video -- at any point in time given only the past frames. For \methodName\ to be causal, the base tracking model needs to be causal. For example, SAM~2 and SAM~2.1 process frames in a streaming fashion, and are, thus, causal. Given this, there are two approaches to allow for causality. 

In the first approach, the global count is provided with a lag of $w$ frames. Stage~1 of \methodName\ is applied to frames $1\cdots t$, and Stages 2 and 3 are applied to frames $1\cdots t - w$, producing a global count at $t - w$. The $w$-frame lag is to allow for filtering false positives in Stage~2. In practice, this lag corresponds to about 1 second.

In the second approach, the global count is provided immediately. Only Stages 1 and 3 of \methodName\ are applied to frames $1\cdots t$, and Stage~2 is skipped. This represents a tradeoff between causality and accuracy as although the global count will be provided more quickly, more false positives will be included in it.

\subsection{Object Quotas of Counters \& Detectors}
Here we briefly discuss how to overcome object quotas of counting and detection models. Detection models based on finite numbers of region proposals, such as Grounding DINO~\cite{liu2023grounding} and OWLv2~\cite{owlv2}, have quotas on the number of objects they can detect in a single frame. This is also true for counting models based on these detectors~\cite{countgd, groundingrec}. To overcome these limits, prior methods~\cite{countgd} propose cropping the image into smaller pieces and summing the counts for each crop while averaging the counts along the overlapping crop boundaries. The issue with these approaches is that averaging overlapping boundaries does not provide bounding boxes for these regions. 

There are ways to address this. For example, predicted boxes with IoUs greater than 0.5 from different crops in overlapping regions can be assigned to one object with a bounding box formed by either averaging the coordinates of the boxes from the different crops, or choosing the box with the highest confidence score. Methods like this provide boxes for objects on the overlapping regions. Additionally, the crop size can be determined by first estimating the object size by averaging the sizes of the predicted bounding boxes and then choosing $crop\_size \ll obj\_size \times quota$. This approximately ensures that the number of objects in each crop is below the model's quota. Future work will extend counting and detection models to count numbers of objects above their quotas using such techniques and plugging these models into the \methodName\ framework.

\subsection{Limitations}
Because \methodName\ builds on a segmentation and tracking model~\cite{ravi2024sam2, yang2024samurai}, it also suffers from some of its shortcomings. For example, the objects in SAM~2 are not `aware' of each other. This means that SAM~2 cannot predict the mask for one object by leveraging information from the predictions for other objects. As a result, object masks can overlap with each other, even if each object has a distinct boundary. Furthermore, the efficiency of the counting and tracking models govern the efficiency of \methodName. Batch processing objects and frames can improve the speed and memory consumption of the model. Since \methodName\ is a general framework, faster and smaller counters and trackers can be swapped in.

Another limitation is that while the video-based MAE and RMSE metrics proposed incorporate re-identification and matching, they do not incorporate identity switching. This is because accounting for identity swaps is not significant for calculating an accurate count. If two objects swap identities, the ground truth count does not change. While the scope of this paper focuses on counting, future work could investigate how well \methodName\ performs as a tracker by evaluating its identity consistency and trajectory accuracy.

\begin{figure}[h]
  \centering
  \includegraphics[width=\linewidth]{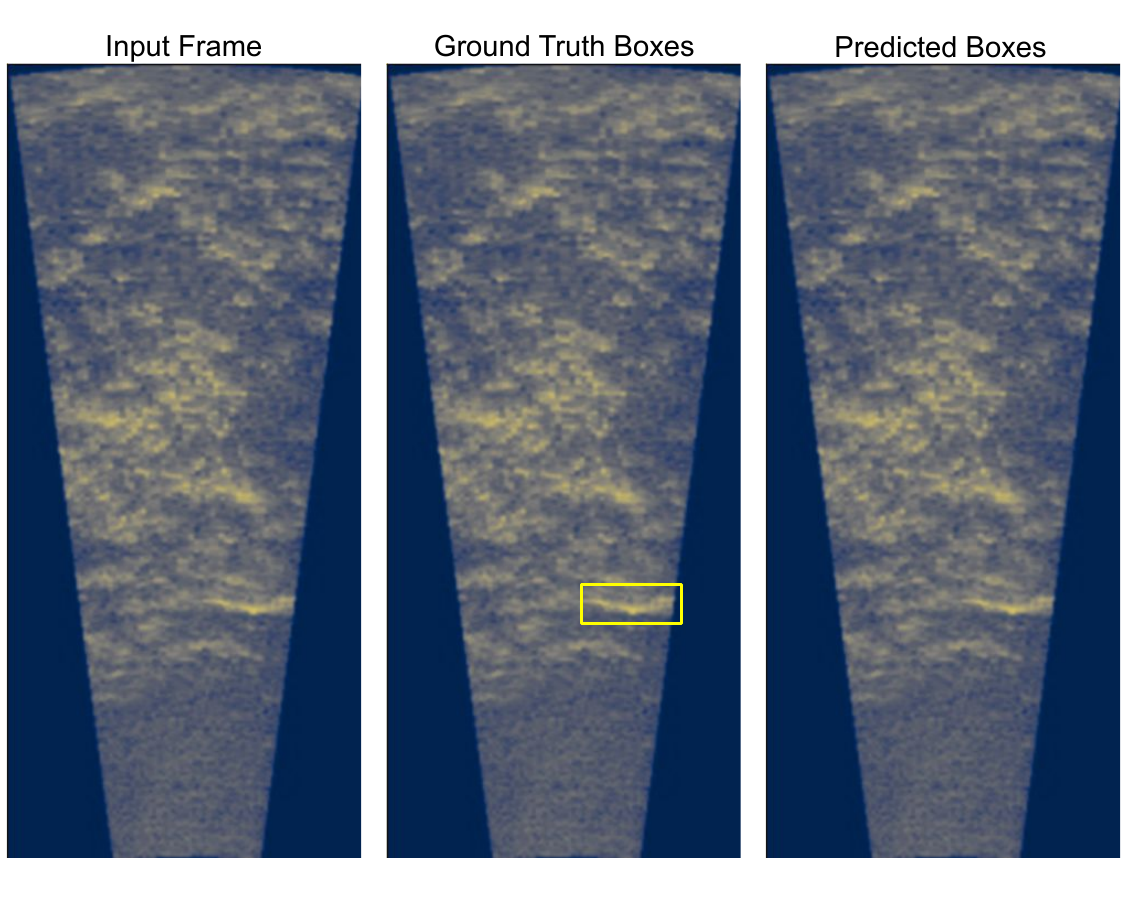}
   \caption{\countgdbox\ struggles to identify the fish in the video frame given the text ``fish,'' since the fish is only clearly visible and distinct from the background when analyzing motion patterns in the water across the video.}
\label{fig:cfc}
\end{figure}

Finally, \methodName\ performs detection at the frame level in Stage~1, and therefore this stage does not use temporal cues. This becomes a limitation in scenarios where objects are not visually distinguishable in individual frames but are only discernible through their motion over time. For example, in the Caltech Fish Counting Dataset~\cite{cfc}, many fish are nearly indistinguishable from the background due to camouflage and low contrast. They become detectable only by observing their movement across sequential frames. As shown in \cref{fig:cfc}, \countgdbox\ fails to localize the fish in Stage~1 when provided only a single frame and the text ``fish.'' Therefore, \methodName\ is unable to count the fish in later stages.

\end{document}